\documentclass[conference]{IEEEtran}
\IEEEoverridecommandlockouts

\usepackage{cite}
\usepackage{amsmath,amssymb}
\usepackage{graphicx}
\usepackage{booktabs}
\usepackage{hyperref}
\usepackage{listings}
\usepackage{xcolor}

\lstdefinelanguage{json}{
    basicstyle=\ttfamily\footnotesize,
    numbers=none,
    stepnumber=1,
    numbersep=8pt,
    showstringspaces=false,
    breaklines=true,
    frame=single,
    literate=
     *{0}{{{\color{black}0}}}{1}
      {1}{{{\color{black}1}}}{1}
      {2}{{{\color{black}2}}}{1}
      {3}{{{\color{black}3}}}{1}
      {4}{{{\color{black}4}}}{1}
      {5}{{{\color{black}5}}}{1}
      {6}{{{\color{black}6}}}{1}
      {7}{{{\color{black}7}}}{1}
      {8}{{{\color{black}8}}}{1}
      {9}{{{\color{black}9}}}{1}
      {:}{{{\color{black}{:}}}}{1}
      {,}{{{\color{black}{,}}}}{1}
      {\{}{{{\color{black}{\{}}}}{1}
      {\}}{{{\color{black}{\}}}}}{1}
      {[}{{{\color{black}{[}}}}{1}
      {]}{{{\color{black}{]}}}}{1},
}

\begin{document}

\title{From World-Gen to Quest-Line: A Dependency-Driven Prompt Pipeline for Coherent RPG Generation}

\author{
\IEEEauthorblockN{Dominik Borawski, Marta Szulc, Robert Chudy, Małgorzata Giedrowicz, Piotr Mironowicz}
\IEEEauthorblockA{
Gdańsk University of Technology \\
Corresponding author: borawskid@icloud.com
}
}

\maketitle

\begin{abstract}

Large Language Models (LLMs) have shown considerable promise for narrative content generation, yet their application to complex, multi-layered role-playing game (RPG) worlds remains limited by issues of coherence, controllability, and structural consistency. This paper investigates a dependency-aware, multi-stage prompt pipeline for procedural RPG content generation that explicitly models narrative dependencies through structured intermediate representations.

The proposed approach decomposes RPG content generation into a sequence of stages: world construction, non-player character generation, player character generation, campaign-level quest planning, and quest expansion, where each stage conditions on the structured JSON outputs of all preceding stages. By enforcing predefined schemas and explicit inter-stage data flow, the pipeline aims to mitigate narrative drift, reduce hallucinations, and enable scalable generation of interconnected narrative artifacts.

The system is evaluated through a qualitative, human-centered assessment across multiple independent pipeline runs. Generated artifacts are analyzed using predefined criteria covering structural completeness, internal consistency, narrative coherence, diversity, and actionability. Results indicate that the proposed pipeline consistently produces logically and structurally valid RPG content across all stages, with no significant degradation in quality as narrative complexity increases. In particular, separating campaign-level planning from quest-level elaboration supports both global narrative order and detailed local storytelling.

The findings suggest that dependency-aware prompt pipelines with structured intermediate representations constitute an effective design pattern for LLM-driven procedural content generation. Beyond RPGs, this approach offers broader applicability to generative systems requiring sequential reasoning over evolving contextual states.

\end{abstract}

\begin{IEEEkeywords}
Large Language Models, Procedural Content Generation, Role-Playing Games, Prompt Engineering, Narrative Generation
\end{IEEEkeywords}

\section{Introduction}
Procedural Content Generation (PCG) has become a fundamental technique in modern role-playing games (RPGs), enabling scalable world-building, increased replayability, and reduced manual authoring costs. Prior research demonstrates that PCG is commonly applied to maps \cite{dungeon_pcg_survey}, quests \cite{de2022procedural} \cite{conan_quest_gen}, and narrative structures \cite{pangia}, primarily with the goal of increasing content diversity rather than narrative depth \cite{hendrikx_pcgsurvey}. As a result, many PCG approaches emphasize variability and coverage over long-term narrative coherence.

Recent advances in Natural Language Processing (NLP), and in particular Large Language Models (LLMs), have introduced new opportunities for narrative-driven content generation in games~\cite{nasir2024word2world,alharthi2025generative}. Multiple studies report promising results in using language models for dialogue generation, quest descriptions, and interactive storytelling \cite{gpt_games_review, rpg_quests_gpt}. Recent survey work further confirms a rapid increase in the adoption of LLMs within procedural content generation research, particularly for narrative and scenario-driven game content, while also highlighting unresolved challenges related to controllability, consistency, and integration with existing PCG pipelines \cite{pcg_llm_survey_2024}. These approaches highlight the expressive capabilities of LLMs and their potential to support creative game design workflows, especially in narrative-heavy genres such as RPGs.

Many PCG systems integrating NLP remain proof-of-concept and focus on isolated content types such as dialogue or single quests, rather than coherent multi-layered RPG worlds \cite{questville}. Furthermore, several studies note that LLM-based generation often relies on single-prompt interactions, which can lead to internal inconsistencies, shallow narrative logic, and limited integration with structured game systems \cite{practical_pcg_llm}. These limitations reduce the practical applicability of LLM-based PCG in production-level RPG pipelines.

While a growing body of work explores the use of procedural content generation and NLP for RPG content, existing approaches typically focus on isolated artifacts such as individual quests, dialogue systems, or world descriptions. Systems such as QuestVille~\cite{questville} and PANGeA~\cite{pangia} demonstrate the feasibility of LLM-based narrative generation but do not explicitly enforce sequential dependencies across multiple layers of RPG content. Planning-based approaches, including CONAN~\cite{conan_quest_gen} and other automated planning frameworks~\cite{de2022procedural}, provide strong guarantees of local narrative coherence but rely on handcrafted domain models and do not integrate world, character, and quest generation within a unified pipeline. To the best of our knowledge, no existing framework combines hierarchical, dependency-aware generation of interconnected RPG artifacts with structured, reusable outputs within a single pipeline suitable for direct integration into game development workflows.

To address this gap, this paper examines a multi-stage, dependency-aware prompt pipeline for generating structured RPG content using an LLM. The proposed approach sequentially generates a game world, a player character, non-player characters, and quest structures, with each generation stage explicitly conditioned on the structured outputs of all preceding stages. By enforcing structured JSON outputs at each stage, the pipeline aims to support consistency, traceability, and practical reusability of generated content.

Specifically, this work investigates whether hierarchical, dependency-aware prompt conditioning improves narrative coherence and structural consistency compared to single-stage or single-prompt generation approaches.

The main contributions of this paper are as follows:
\begin{itemize}
    \item The design of a hierarchical prompt pipeline for dependency-aware RPG content generation.
    \item The use of JSON-constrained outputs to ensure structural validity and practical reusability.
    \item A human-based qualitative evaluation of generated RPG artifacts across multiple content categories.
\end{itemize}

\section{Related Work}

Research on Procedural Content Generation (PCG) has a long tradition in game development, with early work focusing primarily on the generation of maps, levels, and rule-based content to increase replayability and reduce manual authoring effort. More recent work has explored learning-based approaches to PCG, including reinforcement learning and adversarial training setups, where content generators are optimized jointly with solving agents to improve controllability and generalization \cite{arlpcg}. Survey studies indicate that PCG techniques are most commonly applied to structural game elements such as environments, quests, and progression systems, with an emphasis on content diversity rather than deep narrative coherence \cite{togelius_pcgsurvey}. Early comprehensive surveys of PCG further formalized this view by organizing game content into layered taxonomies and highlighting the limited maturity of procedural methods at higher abstraction levels, such as scenarios and game design, compared to low-level structural content \cite{hendrikx_pcgsurvey}. Search-based PCG introduced formal optimization frameworks for PCG by framing content generation as a search problem over explicit representations and fitness functions, thereby enabling controllable generation through objective design rather than handcrafted rules \cite{togelius_pcgsurvey}. Earlier surveys focusing on dungeon and level generation emphasize that the primary limitation of many PCG systems is not expressive power but insufficient designer control, particularly with respect to gameplay semantics, pacing, and progression rather than narrative structure itself \cite{dungeon_pcg_survey}. Experience-driven approaches have further argued for adapting generated content based on player experience models; however, such frameworks largely operate at the level of gameplay parameters and affective response rather than explicit narrative structure or inter-content dependencies \cite{edpcg}. In the context of PCG via Machine Learning (PCGML), prior work has explored a wide range of data representations and learning paradigms for generating functional game content; these approaches, however, primarily target isolated artifacts and do not address multi-layered narrative dependency management \cite{pcgml_survey}. Within RPGs, PCG has been successfully used to automate quest generation and world construction, yet many such approaches rely on handcrafted templates or rule-based logic, limiting their adaptability and narrative flexibility.

More recent work explores the integration of NLP techniques into PCG pipelines to support narrative-driven content generation. Several studies demonstrate the feasibility of using language models for dialogue generation, quest descriptions, and interactive storytelling in games \cite{gpt_games_review, rpg_quests_gpt}. These approaches highlight the expressive capabilities of LLMs and their potential to support creative game design workflows. However, many existing implementations focus on isolated content types, such as single quests or dialogue systems, and are typically evaluated as proof-of-concept prototypes rather than as components of fully integrated RPG content pipelines.

Quest generation represents one of the most actively explored intersections of PCG and NLP in RPG research. Systems such as QuestVille employ NLP techniques to generate quests that are thematically consistent with a game world and player context \cite{questville}. Similarly, PANGeA proposes a narrative generation framework for turn-based RPGs using generative AI techniques \cite{pangia}. While these systems demonstrate promising results in generating coherent individual narrative artifacts, they generally do not address the sequential dependency between multiple layers of RPG content, such as worlds, player characters, non-player characters (NPCs), and quests. As a result, narrative consistency across content layers is typically not enforced explicitly. Beyond NLP-based approaches, prior work has explored the procedural generation of quests using formal planning and evolutionary techniques. Lima et al. propose methods that combine automated planning with genetic algorithms to generate coherent quest structures guided by predefined narrative arcs. While such approaches provide strong guarantees of logical consistency at the quest level, they rely on handcrafted domain models and do not address the integration of quest generation within multi-stage, world-conditioned content pipelines \cite{lima_branching_quests}. Earlier work on procedural quest generation has extensively explored planning-based narrative systems. The CONAN engine generates quests as plans constructed by intentional NPCs based on an explicit representation of the world state, character preferences, and action preconditions. By relying on symbolic planning and handcrafted domain models, such systems provide strong guarantees of logical coherence and character intentionality at the quest level. At the same time, they require substantial manual authoring effort and do not address the integration of quest generation within hierarchical, multi-stage pipelines that condition higher-level narrative elements such as worlds and characters prior to quest construction \cite{conan_quest_gen}.

Recent studies also examine the use of LLMs for practical PCG applications. Khalifa and Togelius highlight the potential of LLMs to simplify content authoring pipelines but note significant challenges related to controllability, consistency, and integration with structured game systems \cite{practical_pcg_llm}. In particular, single-prompt generation approaches often struggle to maintain internal coherence across complex narrative structures, especially when multiple interconnected entities are generated independently. This limitation reduces their suitability for large-scale RPG environments that require tightly coupled narrative elements. Related work on LLM-driven game generation further demonstrates that, while LLMs can synthesize complete game descriptions including rules and levels, they frequently struggle with logical correctness, hallucinations, and adherence to formal structural constraints, underscoring the need for explicit representations and strong structural guidance in generative pipelines \cite{hu_llm_game_generation}.

In contrast to prior work, this paper focuses on a hierarchical, dependency-aware prompt pipeline that sequentially generates interconnected RPG artifacts in a structured JSON format. By explicitly conditioning each generation stage on the structured outputs of all preceding stages, the proposed approach addresses limitations identified in existing PCG and NLP-based systems related to narrative coherence, structural validity, and practical reusability within game development pipelines.

Rather than relying on implicit narrative context encoded in natural language prompts, these limitations motivate the need for a structured, dependency-aware generation framework that explicitly enforces inter-stage dependencies through structured data conditioning, as described in the following section.

\section{System Architecture}

\begin{figure}[t]
    \centering
    \includegraphics[width=\linewidth]{pipeline_diagram.jpg}
    \caption{Dependency-aware multi-stage prompt pipeline for structured RPG content generation. Each generation stage conditions on the complete set of structured JSON outputs produced by all preceding stages.}
    \label{fig:pipeline}
\end{figure}

\subsection{Overview}
The proposed system is designed as a sequential, dependency-aware content generation pipeline for RPGs, as illustrated in Fig.~\ref{fig:pipeline}. The pipeline decomposes RPG content generation into a predefined sequence of stages, each responsible for producing a specific category of game components.

The architecture emphasizes modularity and explicit data dependency management. Each generation stage operates as an independent module with a well-defined input and output interface, while simultaneously remaining contextually grounded in the structured outputs produced by preceding stages. This design enables the inspection of individual stages in isolation while preserving global narrative coherence across the pipeline.

The design of the pipeline is explicitly application-driven rather than benchmark-driven. Following recent recommendations for the evaluation of foundation models in applied settings, the system is optimized for controllability, structural validity, and narrative usability within RPG development workflows, rather than for performance on isolated text generation benchmarks \cite{yuan_science_evaluation}.

\subsection{Core Components}
The system consists of the following core components:

\begin{itemize}
    \item \textbf{Pipeline Controller}: A Python-based orchestration module responsible for executing generation stages in a predefined order, enforcing inter-stage dependencies, and managing the flow of structured data between stages.
    \item \textbf{Large Language Model}: A general-purpose LLM accessed via a commercial LLM API and used without task-specific fine-tuning. The model generates narrative content exclusively based on structured prompts provided by the pipeline.
    \item \textbf{Prompt Templates}: Parameterized prompt definitions that incorporate structured context derived from previous stages and constrain model outputs to predefined JSON schemas.
    \item \textbf{Structured Data Store}: A storage layer for persisting intermediate and final JSON artifacts generated at each stage, enabling reuse, inspection, and validation of generated content.
    \item \textbf{JSON Parser and Validation}: A module that verifies model outputs against predefined templates. If the parser fails during foundational stages (such as World, NPC, or Player generation), the pipeline execution halts, as subsequent stages strictly depend on this structured data. However, parser failures occur most frequently during the final extended quest generation phase. In all cases of failure, the raw text outputs are preserved prior to validation, allowing designers to manually inspect the .txt files and recover the generated narrative content.
\end{itemize}

\subsection{Data Flow and Dependencies}
Content generation follows a strictly sequential and controlled process. The first stage generates a high-level representation of the game world, which establishes the foundational narrative and structural context for all subsequent stages. Based on this world representation, NPCs are generated, followed by a player. Quest generation then utilizes the accumulated structured context of the world, player character, and NPCs to produce narrative objectives and interactions.

Each generation stage is conditioned on the complete set of structured outputs produced by all preceding stages (see examples in Appendix~\ref{app:json_outputs}). This explicit conditioning enforces dependency awareness and reduces the likelihood of inconsistencies between newly generated artifacts and previously established narrative elements.

\subsection{Implementation Context}
The system is implemented in Python and communicates with the OpenAI API through multiple sequential requests issued by the pipeline controller. No model fine-tuning is applied; all behavioral control is achieved through prompt design, structured conditioning, and output constraints. Model parameters are held constant across experiments to ensure comparability of generated outputs and to better observe the effects of the proposed pipeline structure.

The architecture is designed to be extensible. Additional generation stages, validation mechanisms, or post-processing components can be integrated without modifying the core pipeline logic, provided that they adhere to the same structured input-output conventions.

%---
% PROMPT PIPELINE DESIGN
%---

\section{Prompt Pipeline Design}
\label{sec:prompt_pipeline_design}

This section describes the design principles and structure of the multi-stage prompt pipeline used for dependency-aware RPG content generation. The pipeline is composed of a sequence of generation stages, each defined by a dedicated prompt template, an explicitly constructed input context, and a predefined JSON output schema.

At each stage, the LLM is guided through a combination of system-level instructions and task-specific user prompts. System prompts define the global role, style, and output constraints of the model, while user prompts specify the content to be generated and the expected JSON structure. This separation allows for consistent behavioral control across stages while preserving flexibility in content specification.

A key design principle of the pipeline is explicit inter-stage dependency management. Rather than relying on implicit narrative continuity encoded in natural language alone, each stage conditions its generation on structured outputs produced by all preceding stages. These outputs are injected into subsequent prompts as serialized JSON context, ensuring that newly generated artifacts remain consistent with previously established narrative elements.

All generation stages enforce strict output constraints by requiring responses to conform to predefined JSON schemas. This approach serves two purposes: it reduces ambiguity in model outputs and enables deterministic parsing, validation, and reuse of generated artifacts in downstream stages or external systems. Together, these design choices support a controlled yet expressive generation process suitable for multi-layered RPG content creation.

\subsection{World Generation}

The world generation stage constitutes the initial and foundational step of the prompt pipeline. Its objective is to establish a high-level representation of the game world that defines the narrative scope and structural context for all subsequent generation stages. The generated world description includes core elements such as the main city, surrounding locations, key buildings, and political or factional structures.

As the first stage of the pipeline, world generation is not conditioned on any prior narrative artifacts. Instead, the prompt relies exclusively on high-level design intent specified by the user, such as genre or thematic preferences, allowing for maximal creative freedom at the global narrative level. This design choice ensures that downstream stages inherit a coherent and well-defined world context without constraining the initial creative space.

The generation process is guided by a system prompt that establishes the model’s role as a fantasy worldbuilder and enforces strict output constraints. The accompanying user prompt specifies the required world components and their expected organization within a predefined JSON structure. The model is instructed to return only valid JSON compliant with standard parsing rules, with no additional explanatory text.

\subsection{Non-Player Character Generation}

The NPC generation stage is responsible for creating a population of secondary characters that populate the game world and support narrative interactions. These characters serve as quest givers, allies, adversaries, and social connections that shape the player’s experience.

NPCs are generated prior to the player character in order to establish a populated social and political environment into which the player is subsequently introduced.

NPC generation is conditioned on the structured world representation produced in the initial stage of the pipeline. World-level elements such as locations, factions, and thematic context are injected into the prompt as structured input, ensuring that generated characters align with the setting’s lore and political structure. Unlike the player character stage, NPC generation produces multiple character instances simultaneously within a single generation step.

Each NPC is generated according to a predefined JSON schema, illustrated in Listing~\ref{lst:npc}, that encodes identity attributes, personality traits, skills, motivations, secrets, and explicit social relationships with other NPCs. Relationships are represented as structured links between characters rather than as free-form narrative descriptions, enabling the construction of an explicit relational graph across the NPC population.

By enforcing relational structure at the NPC level, the pipeline introduces an additional layer of narrative dependency that can be exploited in downstream stages. Subsequent quest generation stages can reference these relationships explicitly, supporting the creation of conflicts, alliances, and narrative arcs grounded in the existing social structure of the game world.

\subsection{Player Character Generation}

The player character generation stage produces a structured representation of the main hero controlled by the player. Its purpose is to introduce a player-centric perspective into an already established narrative and social environment.

This stage is explicitly conditioned on the structured world representation produced in the world generation stage. In addition, a subset of previously generated NPCs is incorporated into the input context to ground the hero within existing social relationships. By conditioning the prompt on both world-level context and selected NPC references, the pipeline encourages the generation of a protagonist whose history, motivations, and affiliations are narratively consistent with the established setting.

By generating the player character after world generation and the creation of NPCs, the pipeline ensures that the protagonist is not an isolated artifact but an integrated element of the broader narrative ecosystem.

\subsection{Quest Generation}

The quest generation stage is responsible for producing a coherent set of narrative objectives that together form the main storyline of the game. Unlike earlier stages, which focus on individual entities, this stage operates at the campaign level, generating multiple interrelated quests within a single generation step.

Quest generation is explicitly conditioned on the complete structured context accumulated in previous stages, including the game world, the player character, and the population of NPCs. This context is provided to the model as serialized JSON input, allowing quests to reference established locations, factions, characters, and relationships in a consistent manner. As a result, generated quests are grounded in the existing narrative ecosystem rather than created as isolated narrative fragments.

The prompt for this stage enforces a structured JSON schema, as shown in Listing~\ref{lst:quest}, in which each quest includes identifiers, narrative descriptions, objectives, quest-giving characters, rewards, and explicit connections to other quests. Inter-quest dependencies are encoded through structured references, enabling the representation of narrative progression and causal relationships across the campaign. All quests are generated simultaneously within a single model invocation, which promotes global narrative coherence and reduces contradictions between individual quest definitions.

By treating quest generation as a campaign-level task rather than a sequence of independent prompts, the pipeline establishes a high-level narrative backbone that can be further refined in subsequent stages.

\subsection{Extended Quest Generation}

The extended quest generation stage refines the high-level campaign structure produced in the previous stage by expanding individual quests into detailed, multi-step narrative experiences. In contrast to the campaign-level quest generation process, this stage operates at the granularity of single quests and is executed iteratively for each quest in the generated storyline.

Each extended quest generation step conditions on the full structured context of the game world, player character, NPC population, and the specific quest selected for expansion. This design allows the model to elaborate narrative details such as dialogue, player choices, consequences, and branching outcomes while preserving consistency with both the global narrative context and the original quest intent.

The output of this stage adheres to a more expressive JSON schema (see Listing~\ref{lst:extended_quest}) that augments the base quest structure with rich narrative components, including extended descriptions, multi-turn dialogues, explicit player decision points, and reward elaborations. By isolating quest expansion into separate generation calls, the pipeline enables fine-grained control over narrative depth and allows individual quests to be refined, regenerated, or evaluated independently.

This two-tiered approach to quest generation separates global narrative planning from local narrative elaboration. As a result, the pipeline balances campaign-level coherence with detailed, quest-specific storytelling, supporting scalable narrative generation without sacrificing structural consistency.

\section{Experimental Setup}

\subsection{Model and API Configuration}

All experiments were conducted using a general-purpose LLM accessed via the OpenAI API. The GPT-5 model was employed in a zero-shot setting without task-specific fine-tuning or additional training.

All interactions with the model were performed through a chat-based completion interface using system and user role messages. Model parameters were held constant across all generation stages to ensure comparability of outputs. In particular, the temperature parameter was fixed to a value of 1.0 for world generation, player character generation, NPC generation, quest generation, and extended quest generation.

A uniform maximum token limit was applied across all generation stages to prevent variability arising from differing output capacities. This configuration ensured that observed differences in generated content were attributable to prompt structure and dependency-aware conditioning rather than to changes in model configuration.

\subsection{Pipeline Configuration and Generation Procedure}

The prompt pipeline is configured as a strictly sequential, multi-stage generation process in which each stage incrementally expands the narrative universe while remaining explicitly dependent on all previously generated content. The ordering of stages is intentionally designed to progress from global, high-level representations toward increasingly detailed and localized narrative artifacts, ensuring that narrative elements do not emerge in isolation but instead remain embedded within a shared and evolving fictional context.

Recall that the pipeline consists of five stages executed in a fixed order: world generation, NPCs generation, player character generation, quest generation, and extended quest generation. Each stage grounds newly generated artifacts in the structured outputs of all preceding stages, as discussed in Sec.~\ref{sec:prompt_pipeline_design}, allowing later generation steps to remain responsive to earlier design decisions such as world lore, character attributes, social relationships, and narrative structure.

Based on these structural constraints and predefined JSON schemas, a single experimental run is defined as a complete execution of the prompt pipeline, starting from world generation and ending with extended quest generation. Each run produces a full set of structured RPG artifacts, including a world description, a population of NPCs, a player character profile, a campaign-level quest set, and expanded versions of individual quests (refer to Appendix~\ref{app:json_outputs} for schema excerpts).

Pipeline execution follows a strictly sequential procedure. Each generation stage is executed as an independent model invocation using a dedicated prompt template and conditioned exclusively on the structured JSON outputs produced by all preceding stages. No conversational memory or hidden state is preserved between stages beyond the explicitly provided structured context.

All intermediate and final outputs are persisted as structured JSON artifacts immediately after generation. These artifacts serve both as input for subsequent pipeline stages and as the basis for later evaluation, enabling traceability across stages and ensuring that all downstream generation steps operate on identical, unmodified data.

After completion of the textual generation stages, an optional visualization process is applied to selected artifacts to produce illustrative assets. This post-processing step does not modify any textual content and is not considered part of the core experimental results.

\subsection{LLM Configuration and Generation Parameters}

All content generation stages in the proposed pipeline were implemented using the GPT-5 LLM. A single, fixed model configuration was used across all stages and experimental runs to ensure consistency of the generated artifacts.

Each generation stage was allocated a maximum output budget of 32,768 tokens. This relatively large token limit was chosen to allow the model to produce detailed, self-consistent narrative structures without being forced into premature truncation, particularly at higher levels of abstraction such as world and quest generation.

Sampling was performed using a uniform temperature setting of 1.0 across all stages of the pipeline, including world generation, player character creation, NPC generation, and quest construction. No stage-specific temperature tuning was applied. This design choice reflects an intentional separation between deterministic structural control, enforced through explicit JSON schemas and dependency propagation, and stochastic narrative content generation, which remains unconstrained at the sampling level.

All experiments were conducted using the same model version and configuration over a fixed experimental period.

\subsection{Experimental Scope}

The experimental evaluation was conducted over multiple independent executions of the prompt pipeline. In total, twenty complete pipeline runs were performed. Each run was executed from scratch and did not reuse any artifacts generated in previous runs.

A single pipeline run produced a fixed set of structured RPG artifacts. Specifically, each run generated one game world, one player character, ten NPCs, ten campaign-level quests, and ten extended quest instances derived from the generated campaign structure. As a result, the full experimental dataset comprised 20 distinct worlds, twenty player characters, two hundred NPCs, two hundred campaign-level quests, and two hundred extended quests.

All runs were executed using identical pipeline configuration, prompt templates, and model parameters. No adaptive tuning or manual intervention was applied between runs. This setup was chosen to assess the consistency and robustness of the proposed pipeline under repeated executions while allowing for natural variability in narrative content.

The generated artifacts formed the basis for the qualitative evaluation described in the following section.

In line with recent recommendations on evaluating foundation model–based systems, the unit of evaluation in this study is not an individual model output but a complete pipeline execution \cite{yuan_science_evaluation}. Each pipeline run represents a coherent deployment instance of the proposed system, producing an interconnected set of narrative artifacts whose quality can only be meaningfully assessed in aggregate.

\subsection{Reproducibility and Controls}

All experiments were conducted using fixed prompt templates, model parameters, and pipeline configuration across all runs. No manual edits or post-processing modifications were applied to any generated artifacts at any stage of the pipeline. Each generation stage operated solely on structured outputs produced by preceding stages, ensuring consistent dependency-aware conditioning across runs.

All intermediate and final artifacts were stored as structured JSON files immediately after generation, enabling inspection and traceability of the complete generation process. While the underlying language model exhibits inherent stochasticity, repeated executions were performed under identical conditions to assess qualitative consistency rather than deterministic reproducibility.

\subsection{Computational Environment}

The pipeline was implemented in Python~3.12 and executed on standard desktop and laptop systems running Windows and macOS operating systems. Interaction with LLMs was performed via a commercial API using the \texttt{openai} Python library (version~$\geq$~1.0.0). Environment variable management was handled using the \texttt{python-dotenv} library (version~$\geq$~1.0.0).

Structured data processing and validation relied on standard Python tooling, with relational structures represented and manipulated using the \texttt{networkx} library (version~$\sim$~3.5). No specialized hardware or GPU acceleration was required, as all model inference was performed externally via API calls.

\section{Evaluation Protocol}

This section describes the qualitative evaluation protocol used to assess the outputs generated by the proposed dependency-aware prompt pipeline. Given the narrative and creative nature of RPG content, the evaluation focuses on qualitative aspects that reflect narrative coherence, structural validity, and practical usability rather than on automatic text similarity metrics.

\subsection{Evaluation Objectives}

The primary objective of the evaluation is to assess whether the proposed multi-stage, dependency-aware prompt pipeline produces RPG content that is coherent, internally consistent, and suitable for direct use in game development contexts. In particular, the evaluation aims to determine whether explicit inter-stage conditioning and structured JSON outputs improve the quality of generated artifacts compared to isolated or single-prompt generation approaches.

A secondary objective is to examine the robustness of the pipeline across multiple independent runs. By evaluating artifacts generated under identical configuration settings, the study seeks to identify recurring strengths and limitations of the approach, as well as common failure modes related to narrative structure or consistency.

Rather than measuring linguistic similarity or stylistic novelty, the evaluation emphasizes properties that are directly relevant to RPG design workflows, such as narrative plausibility, cross-entity consistency, and actionability of generated content. This focus reflects the intended application of the pipeline as a content generation tool for structured, narrative-driven game systems.

\subsection{Evaluation Criteria}

The evaluation is based on a set of qualitative criteria designed to capture key dimensions of quality in structured RPG content. These criteria were selected to reflect both narrative and technical considerations relevant to PCG and game development practice.

\textbf{Structural Completeness} assesses whether each generated artifact conforms to its predefined JSON schema and contains all required fields in a meaningful and usable form. This criterion reflects the importance of structural validity for downstream processing and direct integration into game engines or content pipelines.

\textbf{Internal Consistency} evaluates the degree to which generated artifacts remain logically consistent with previously established narrative elements, such as world lore, character attributes, and inter-character relationships. This criterion directly targets the effectiveness of dependency-aware conditioning in reducing contradictions across content layers.

\textbf{Narrative Coherence} measures the extent to which generated content forms a plausible and comprehensible narrative within the fictional setting. This includes the clarity of motivations, causal relationships between events, and alignment between characters, locations, and quests.

\textbf{Diversity} captures the variability and distinctiveness of generated artifacts within a single pipeline run. For example, this criterion assesses whether NPCs and quests exhibit meaningful differences in roles, personalities, and objectives rather than collapsing into repetitive patterns.

\textbf{Actionability} evaluates whether the generated content can be readily used or adapted by a game designer without substantial manual rewriting. This includes the clarity of quest objectives, the usefulness of character descriptions, and the practical applicability of generated structures within a game design context.

These criteria were chosen in place of automatic text-based metrics because the generated artifacts are highly structured, narrative-driven, and context-dependent. 
Metrics such as BLEU~\cite{papineni2002bleu} and ROUGE~\cite{lin2004rouge} are known to correlate poorly with human judgments in complex natural language generation (NLG) tasks~\cite{novikova2017we,celikyilmaz2020evaluation}. For structured, context-dependent narrative artifacts, qualitative human-centered evaluation is therefore often recommended~\cite{van2019best}. In the context of foundation models, recent work further emphasizes evaluating systems with respect to task-specific goals and deployment constraints rather than generic linguistic similarity~\cite{bommasani2021opportunities,liang2022holistic}.
%Metrics such as BLEU or ROUGE are poorly suited to capturing narrative quality, structural validity, or design usefulness in this setting. A qualitative, human-centered evaluation was therefore deemed more appropriate for assessing the practical impact of the proposed pipeline. The selected criteria are explicitly aligned with the intended use of the system as a narrative content generation tool for RPG development, reflecting recent recommendations to evaluate foundation models with respect to task-specific goals, deployment constraints, and practical usability rather than generic linguistic similarity \cite{yuan_science_evaluation}.

\subsection{Artifact Sampling}

The evaluation was conducted on a subset of artifacts generated across multiple pipeline runs. Sampling focused on ensuring coverage of all content categories produced by the pipeline, including worlds, player characters, NPCs, campaign-level quests, and extended quests.

World-level artifacts were selected by randomly sampling several complete pipeline runs. For each selected world, the corresponding dependent artifacts generated within the same run (NPCs, player character, quests, and extended quests) were included in the evaluation. This strategy preserved the full dependency structure between artifacts while allowing assessment across multiple independent narrative settings.

The sampling approach was designed to balance coverage and feasibility in the context of a qualitative, human-centered evaluation. Rather than aiming for statistical representativeness, the selected artifacts were intended to provide insight into typical outputs and recurring patterns produced by the proposed pipeline.

\subsection{Evaluation Procedure}

The evaluation was conducted by four evaluators using a structured qualitative assessment protocol applied uniformly across all sampled artifacts. For each selected pipeline run, evaluators independently reviewed the generated world artifact together with all dependent artifacts produced within the same run, including the player character, NPCs, campaign-level quests, and extended quests.

Artifacts were evaluated directly on unedited model outputs, with no manual correction or post-processing, ensuring that assessed artifacts faithfully represent the system’s behavior. Each artifact was assessed independently against the predefined evaluation criteria (Structural Completeness, Internal Consistency, Narrative Coherence, Diversity, and Actionability) using a five-point Likert scale. Scores were recorded per criterion and per content category to enable later aggregation and comparison across stages of the pipeline.

In addition to numerical scores, evaluators recorded qualitative notes capturing representative observations, recurring strengths, and common failure modes. These notes were subsequently used to contextualize numerical results and support the interpretation of results and the discussion of limitations. Summary statistics were computed by aggregating criterion-level scores across evaluators and artifacts within each content category, forming the basis for the reported results.

\subsection{Evaluators Background}

All four evaluators participated in the design and implementation of the proposed pipeline and were therefore familiar with its intended functionality and design goals. They were graduate-level students with backgrounds in computer science or related fields, possessing prior experience as players of both digital and tabletop RPGs. Furthermore, their familiarity with LLMs, prompt engineering, and PCG enabled informed assessment of the generated artifacts with respect to both narrative quality and practical usability.

All evaluated artifacts were generated and assessed in English, which was the evaluators’ working language. No formal calibration session was conducted prior to evaluation; however, all evaluators were provided with the same written evaluation guidelines and criterion definitions to promote consistent interpretation of the scoring scheme.

\subsection{Scoring Scheme}

The five-point Likert scale was designed to capture the perceived quality of generated artifacts along each criterion while remaining simple enough to support consistent application by a single evaluator. Each criterion was scored independently, where 1 indicated very poor performance with severe deficiencies, 3 indicated acceptable or moderate quality, and 5 indicated high-quality output without observable degradation in qualitative assessments as narrative complexity increases. Intermediate values were used to express gradations between these anchor points.

\subsection{Reliability and Bias Considerations}

While the evaluators’ involvement in system development enabled informed expert assessment, it inherently introduces potential subjectivity and confirmation bias. This approach, however, is consistent with exploratory system and design studies, where expert inspection by system authors is commonly employed.

Several measures were taken to mitigate these limitations. As detailed above, evaluations were conducted independently using a fixed set of criteria, and artifacts were sampled from multiple independent pipeline executions to reduce the likelihood that results reflect a single favorable generation instance. Furthermore, criterion-level scores were aggregated across evaluators, and qualitative notes were compared to identify recurring patterns.

However, no formal inter-rater agreement analysis was conducted, which limits the strength of quantitative conclusions drawn from the Likert-scale scores. Despite the mitigation strategies, the reported results should be interpreted as exploratory and indicative rather than definitive. Future work will address these limitations through the inclusion of external evaluators, formal inter-rater reliability analysis, evaluator calibration procedures, and user studies involving players and professional game designers.

\section{Results}

\subsection{Overview of Evaluated Artifacts}

The results reported in this section are based on a qualitative evaluation of artifacts generated across multiple independent executions of the proposed prompt pipeline. In total, twenty complete pipeline runs were considered, each producing a consistent set of structured RPG artifacts.

For each pipeline run, the system generated one game world, one player character, ten NPCs, ten campaign-level quests, and ten extended quests. Evaluation focused on a sampled subset of these artifacts, selected to ensure coverage of all content categories and multiple independent narrative settings.

\subsection{Aggregate Scores by Content Category}

Table~\ref{tab:aggregate_scores} summarizes the aggregated qualitative evaluation results across all sampled artifacts and content categories. Scores represent average values computed from the five-point Likert-scale assessments assigned to each evaluation criterion.

Results are reported separately for each major content category produced by the pipeline: World, NPCs, Player Character, Campaign-Level Quests, and Extended Quests. For each category, mean scores are presented for Structural Completeness, Internal Consistency, Narrative Coherence, Diversity, and Actionability, along with an overall average score computed as the unweighted mean across criteria.

The reported values provide a compact overview of the relative performance of the pipeline across different stages of content generation and serve as the basis for the more detailed, criterion-level analysis presented in subsequent subsections.

\begin{table}[t]
\centering
\caption{Aggregate qualitative scores by content category (five-point Likert scale)}
\label{tab:aggregate_scores}
\resizebox{\columnwidth}{!}{%
\begin{tabular}{@{}lcccccc@{}}
\toprule
Content Category & Struct. & Consist. & Coher. & Divers. & Action. & Avg. \\
\midrule
World              & 5.00 & 4.20 & 4.40 & 4.60 & 4.80 & 4.60 \\
NPCs               & 5.00 & 4.20 & 3.60 & 4.60 & 4.00 & 4.28 \\
Player Character   & 4.80 & 3.80 & 4.00 & 4.60 & 3.80 & 4.20 \\
Quests             & 4.40 & 4.40 & 4.40 & 4.20 & 3.40 & 4.16 \\
Extended Quests    & 3.60 & 4.20 & 4.20 & 4.00 & 4.60 & 4.12 \\
\bottomrule
\end{tabular}%
}
\end{table}

\section{Discussion}

\subsection{Interpretation of Aggregate Results}

The aggregate results presented in Section~V indicate consistently high qualitative scores across all evaluated content categories. Notably, no substantial degradation in quality is observed in later stages of the pipeline, despite the increasing narrative complexity and dependency depth. This suggests that the sequential conditioning strategy effectively preserves contextual information throughout the generation process.

World-level artifacts achieve particularly strong scores, reflecting the stability of the foundational narrative layer. Importantly, subsequent stages - especially quest and extended quest generation - do not exhibit signs of narrative collapse or incoherence, which are commonly reported issues in single-prompt or flat generation approaches. Instead, quests appear to meaningfully build upon previously generated world and character elements, while extended quests further elaborate these structures without introducing contradictions.

These observations suggest that explicitly conditioning later stages on structured outputs from earlier stages enables the pipeline to maintain narrative continuity and structural integrity, even as the scope and detail of generated content increase.

\subsection{Effect of Dependency-Aware Prompt Conditioning}

The primary contribution of the proposed approach lies in its use of dependency-aware prompt conditioning combined with strongly structured inputs and outputs. By enforcing JSON-based schemas at each stage, the pipeline constrains generation in a manner that reduces hallucinations while preserving expressive flexibility.

Structured representations serve as explicit carriers of narrative state, allowing each generation stage to operate on well-defined and reusable contextual artifacts. This design contrasts with natural-language-only prompting, where contextual information is implicitly encoded and prone to drift or loss across longer generations. In the proposed pipeline, structured conditioning enables the model to reason over persistent world, character, and relational data, which is reflected in the consistently high scores for internal consistency and narrative coherence.

An additional benefit of the multi-stage approach is its ability to mitigate token limit constraints. Rather than attempting to generate all content within a single prompt, the pipeline distributes generation across multiple stages, each operating on a manageable context size. This allows individual stages to produce more detailed and structurally rich outputs than would be feasible in a single-pass generation, while maintaining global coherence across the narrative universe.

Overall, the results suggest that dependency-aware, structured prompt pipelines provide a viable mechanism for scaling LLM-based narrative generation beyond isolated artifacts, particularly in domains such as RPGs that require long-term narrative consistency.

\subsection{Campaign-Level versus Quest-Level Generation}

Maintaining coherence across multiple interconnected quests represents a significant challenge for LLM-based narrative generation. When quests are generated independently, there is a high risk of losing global campaign structure, narrative pacing, and inter-quest dependencies. The proposed pipeline addresses this challenge by explicitly separating campaign-level planning from quest-level elaboration.

By first generating a complete set of campaign-level quests within a single generation step, the pipeline establishes a high-level narrative backbone that encodes progression, thematic continuity, and inter-quest relationships. This global structure constrains subsequent generation and reduces the likelihood of contradictions between individual quests. The relatively strong coherence scores observed for both campaign-level and extended quests support this design choice.

Maintaining coherence across multiple interconnected quests represents a significant challenge for LLM-based narrative generation. When quests are generated independently, there is a high risk of losing global campaign structure, narrative pacing, and inter-quest dependencies. The proposed pipeline addresses this challenge by explicitly separating campaign-level planning from quest-level elaboration.

As demonstrated by the relatively strong coherence scores (4.40 for campaign-level quests and 4.20 for extended quests), establishing a high-level narrative backbone prior to local expansion effectively reduces the likelihood of contradictions between individual quests. By constraining subsequent generation with this global structure, the system successfully preserves thematic continuity and inter-quest relationships without the need for manual intervention.

Furthermore, the combination of strong actionability (4.60) and coherence scores for extended quests suggests that this two-tiered approach effectively balances global narrative control with local narrative depth. It enables the generation of richer local narratives—such as multi-turn dialogues and branching interactions—while strictly maintaining alignment with the overarching campaign.

\subsection{Practical Implications for RPG Content Pipelines}

From a practical perspective, the proposed pipeline illustrates how structured representations can function as a lingua franca between LLMs, human designers, and game engines. JSON-based artifacts (such as the extended quest shown in Listing~\ref{lst:extended_quest}) provide a transparent and manipulable interface through which generated content can be inspected, modified, validated, and integrated into downstream systems.

The use of structured outputs enables incremental and potentially unbounded expansion of a narrative universe. Individual elements, such as locations, characters, or quests - can be selectively injected into prompts as contextual building blocks, allowing designers to guide generation while preserving systemic coherence. This modularity supports workflows in which human oversight and automated generation coexist, facilitating rapid prototyping and iterative refinement.

Beyond RPG design, the principles demonstrated by the pipeline may generalize to other domains that require sequential reasoning over evolving contexts. Applications such as interactive storytelling, simulation design, educational content generation, and scenario planning similarly benefit from explicit state representation and dependency-aware generation. In these settings, structured prompt pipelines may provide a scalable alternative to monolithic generation approaches.

\subsection{Limitations and Threats to Validity}

Several limitations should be considered when interpreting the results of this study. First, the evaluation relies on qualitative judgments provided by a single evaluator, introducing subjectivity and limiting inter-rater reliability. While consistent criteria were applied across all artifacts, the findings cannot be assumed to generalize across evaluators with different backgrounds or expectations.

Second, the study does not employ automated quantitative metrics for narrative quality, coherence, or diversity. Although such metrics remain challenging to define for narrative content, their absence restricts the ability to perform large-scale or statistically grounded comparisons. Additionally, no baseline comparison against alternative generation strategies, such as single-prompt or flat generation pipelines, was conducted within the same experimental framework.

Finally, the experiments were performed using a single LLM configuration. Differences in model architecture, training data, or context handling may affect the observed outcomes. Future work should examine the robustness of the proposed pipeline across models and parameter settings.

These limitations reflect broader challenges identified in recent work on the evaluation of foundation models, where qualitative, human-centered assessment remains necessary but inherently constrained by subjectivity, evaluator bias, and scalability concerns \cite{yuan_science_evaluation}. The present study therefore positions its findings as exploratory evidence supporting the proposed design pattern rather than as definitive performance claims.

\subsection{Relation to Prior Work}

Prior research on PCG and narrative systems has explored both rule-based and learning-based approaches to quest and world generation. Recent studies leveraging LLMs demonstrate the feasibility of generating individual narrative artifacts, such as quests or dialogues, but often treat these artifacts in isolation.

In contrast, the proposed pipeline emphasizes explicit dependency management across multiple layers of RPG content. Unlike systems that rely on single-prompt generation or loosely structured textual context, this approach integrates structured representations and sequential conditioning to enforce cross-layer consistency. The findings align with observations in prior work regarding the limitations of unconstrained LLM-based generation, while extending this line of research by demonstrating a practical mechanism for maintaining narrative coherence at scale.

By framing content generation as a staged, dependency-aware process, the pipeline contributes a system-level perspective that complements existing work focused on individual narrative components.

\subsection{Summary of Insights}

Overall, the results suggest that dependency-aware prompt pipelines with structured intermediate representations provide a promising foundation for narrative-driven PCG. The proposed approach demonstrates that LLMs can be used not only to generate isolated narrative artifacts, but also to construct coherent, multi-layered narrative systems when guided by explicit structure and sequential conditioning.

While the findings are exploratory and subject to the limitations discussed above, they indicate that structured prompt pipelines represent a viable direction for future research and practical tool development in RPG content generation and related narrative domains.

\section{Conclusion and Future Work}

This paper investigated the use of a dependency-aware, multi-stage prompt pipeline for structured RPGs content generation using LLMs. The proposed approach decomposes narrative generation into sequential stages that explicitly condition on structured outputs produced in earlier steps, enabling the construction of coherent, multi-layered narrative artifacts.

The qualitative evaluation indicates that the pipeline is capable of generating structurally valid, internally consistent, and actionable RPG content across multiple content categories, including worlds, characters, and quests. Importantly, the results suggest that explicit dependency management and structured intermediate representations help mitigate common issues associated with single-prompt generation, such as narrative drift and loss of global coherence as content complexity increases.

Beyond the specific application to RPGs, the findings highlight the broader potential of structured, dependency-aware prompt pipelines as a design pattern for LLM–driven systems that require sequential reasoning over evolving contexts. By separating global planning from local elaboration and enforcing explicit structure at each stage, such pipelines offer a promising balance between creative flexibility and architectural control.

Future work will address several limitations of the current study. Planned extensions include multi-rater and user-centered evaluations involving game designers and players, the integration of automated validation and analysis techniques, and systematic comparisons against baseline generation strategies. Additionally, future research will explore the applicability of the proposed pipeline across different language models, as well as hybrid approaches that combine prompt-based control with symbolic reasoning or fine-tuning to further enhance consistency and controllability.

Overall, this work contributes a practical and extensible framework for narrative-driven PCG and suggests that dependency-aware prompt pipelines represent a viable direction for both research and tool development in complex generative systems.

\appendices

\section{Repository and Reproducibility Materials}

The complete implementation of the proposed dependency-aware prompt pipeline, together with all generated artifacts referenced in this paper, is publicly available in a dedicated GitHub repository.\footnote{Repository URL: \url{https://github.com/borawskiD/PCG_in_RPG_Systems_Using_LLM}}

The repository contains the full pipeline implementation, prompt templates, JSON schemas, raw language model outputs, and all structured artifacts produced during experimental runs. A fixed commit hash is provided to ensure reproducibility of the reported results.

Due to the volume of generated content, this appendix includes only representative excerpts of structured outputs. The repository provides complete access to all generated artifacts required to reproduce and inspect the pipeline behavior.

\section{Structured JSON Outputs}
\label{app:json_outputs}

This appendix presents representative excerpts of the structured JSON artifacts produced by the proposed pipeline. The examples illustrate schema design, inter-artifact dependencies, and progressive refinement across pipeline stages. Due to the size and number of generated files per execution, only selected excerpts are included here; complete artifacts are provided in the accompanying repository.
Accordingly, the listings below focus on illustrating structural dependencies and data flow between generation stages rather than exhaustively documenting the full JSON schemas.

\subsection{World Representation}

The world generation stage produces a single structured JSON artifact defining the global narrative context referenced by all downstream stages.
Only fields relevant for cross-stage referencing are shown.

\lstset{breakatwhitespace=true}
\begin{lstlisting}[
  basicstyle=\ttfamily\footnotesize,
  breaklines=true,
  frame=single,
  caption={World-level structured representation (excerpt)},
  label={lst:world}
]
{
  "city": "Marinth Vell, the Tideglass Metropolis...",
  "surroundings": [
    {
      "name": "Storm-Kiln Caldera",
      "type": "volcanic caldera and storm-factory",
      "resources": "Storm energy, fulgurites",
      "dependencies": "Storm access required for tideglass production",
      "related_factions": [
        "Glassworkers' Concord",
        "Conclave of Prismancers",
        "House Saltvein"
      ]
    },
    ...
  ],
  "buildings": [
    {
      "name": "Beacon of Myr Sal",
      "type": "living lighthouse",
      "district": "Glassward",
      "controlled_by": "Conclave of Prismancers and Faith of Azerene"
    },
    ...
  ],
  "politics": [
    {
      "name": "Glassworkers' Concord",
      "type": "craft guild",
      "goals": "Maintain control over tideglass craft",
      "rivals": "Conclave of Prismancers"
    },
    ...
  ]
}
\end{lstlisting}

Listings show representative excerpts of generated artifacts; omitted fields and repeated entries are indicated by ellipses for brevity.

\subsection{Non-Player Characters}

Non-player characters are instantiated within the previously generated world and inherit its factions, locations, and political structures as contextual constraints explicitly referenced in their JSON representations. Each NPC is encoded as an individual JSON artifact capturing personality traits, competencies, internal motivations, hidden information, and explicit social relationships.

\begin{lstlisting}[
  caption={Non-player character representation (excerpt)},
  label={lst:npc}
]
{
  "name": "Isha Brinehand",
  "role": "Foreman-Keeper of the Dockers' Syndicate",
  "traits": [
    "charismatic",
    "protective",
    "combative"
  ],
  "skills": [
    "strike organization",
    "crowd speaking",
    "contract bargaining"
  ],
  "flaws": [
    "impulsive",
    "holds grudges"
  ],
  "secrets": [
    "Funded lantern blackouts during labor negotiations",
    "Shelters a deserter Warden under a false docker's name"
  ],
  "relations": [
    { "npc_name": "Sorev Katch", "relation_type": "Conflict" },
    { "npc_name": "Sel Var Saltvein", "relation_type": "Rivalry" },
    { "npc_name": "Nerey Alis", "relation_type": "Cooperation" }
  ]
}
\end{lstlisting}

Fields such as appearance, history, and extended motivations are omitted for brevity; the complete NPC schema and full instances are available in the repository.

\subsection{Player Character}

The player character is generated with explicit references to previously instantiated non-player characters, allowing relationships encoded in the player JSON to directly bind the protagonist to existing social and political actors.

\begin{lstlisting}[
  basicstyle=\ttfamily\footnotesize,
  breaklines=true,
  frame=single,
  caption={Player character representation (excerpt)},
  label={lst:player}
]
{
  "name": "Tairn Latch",
  "class": "WARRIOR",
  "background": "Marsh-born chainhouse rigger...",
  "main_attributes": {
    "strength": 16,
    "constitution": 15,
    "dexterity": 14
  },
  "relationships": {
    "Isha Brinehand": "Mentor",
    "Sel Var Saltvein": "Rival"
  }
}
\end{lstlisting}

The excerpt highlights the structural elements relevant for quest grounding and social interaction; full narrative background and equipment descriptions are provided in the repository.

\subsection{Campaign-Level Quests}

Campaign-level quests are generated in a single stage as a coherent set. Each quest references NPCs, locations, and other quests through explicit identifiers.

At this stage, quests are intentionally encoded as lightweight planning artifacts, omitting narrative detail in order to separate global campaign structure from later, dependency-aware narrative elaboration.
This design prevents premature narrative commitment and enables consistent elaboration across multiple quests in later stages.
The apparent simplicity of campaign-level quests is a deliberate design choice rather than a limitation of the model.
\newpage 

\begin{lstlisting}[
  basicstyle=\ttfamily\footnotesize,
  frame=single,
  caption={Campaign-level quest (excerpt)},
  label={lst:quest}
]
{
  "id": "M1",
  "title": "When the Beacon Keens on a Clear Sky",
  "quest_giver": "Isha Brinehand",
  "objectives": [
    "Stabilize the chainlift at Chainmouth",
    "Meet Captain-Major Sorev Katch"
  ]
}
\end{lstlisting}

Inter-quest links encode campaign-level narrative progression and are preserved during later elaboration.

\subsection{Extended Quest Elaboration}

In the final stage, campaign-level quests are elaborated into extended narrative artifacts that explicitly reference previously generated world entities, non-player characters, and the player character, while preserving original identifiers and dependency structure.

\begin{lstlisting}[
  basicstyle=\ttfamily\footnotesize,
  breaklines=true,
  frame=single,
  caption={Extended quest representation (excerpt)},
  label={lst:extended_quest}
]
{
  "id": "M1",
  "title": "When the Beacon Keens on a Clear Sky",
  "quest_giver": "Isha Brinehand",
  "objectives": [
    "Stabilize the jammed chainlift at Chainmouth",
    "Report findings to Captain-Major Sorev Katch"
  ],
  "dialogue": [
    {
      "speaker": "Isha Brinehand",
      "content": "The Beacon shouldn't sing when the sky is clean..."
    },
    {
      "speaker": "Tairn Latch",
      "content": "I'll climb and listen if the Conclave will let me."
    },
    {
      "speaker": "Sorev Katch",
      "content": "Bring me something I can act on."
    }
  ],
  "connections": {
    "next": ["M2"]
  },
  "rewards": [
    "Dockers' Syndicate hazard pay",
    "River Wardens access pass"
  ]
}
\end{lstlisting}

The complete quest representation, including extended narrative descriptions, branching dialogue, and consequence propagation, is provided in the accompanying repository.

\subsection{Raw and Structured Output Separation}

Raw language model outputs are preserved prior to structural transformation and validation. These unstructured texts are subsequently converted into validated JSON artifacts consumed by downstream generation stages.

Maintaining both representations enables inspection of model behavior, debugging of schema violations, and analysis of narrative drift.

\bibliographystyle{IEEEtran}
\bibliography{references}

@article{togelius_pcgsurvey,
  author  = {J. Togelius and G. N. Yannakakis and K. O. Stanley and C. Browne},
  title   = {Search-Based Procedural Content Generation: A Taxonomy and Survey},
  journal = {IEEE Transactions on Computational Intelligence and AI in Games},
  volume  = {3},
  number  = {3},
  pages   = {172--186},
  year    = {2011},
  doi     = {10.1109/TCIAIG.2011.2148116}
}

@article{questville,
  author  = {E. Soares de Lima and M. M. E. Neggers and B. Feij{o} and M. A. Casanova and A. L. Furtado},
  title   = {QuestVille: Procedural Quest Generation Using NLP Models},
  journal = {Entertainment Computing},
  volume  = {47},
  year    = {2024},
  doi     = {10.1016/j.entcom.2024.100810}
}

@article{gpt_games_review,
  author  = {J. Bergdahl and S. Dahlskog},
  title   = {GPT for Games: A Scoping Review (2020--2023)},
  journal = {IEEE Transactions on Games},
  year    = {2023}
}

@inproceedings{rpg_quests_gpt,
  author    = {X. Peng and J. Quaye and S. Rao and W. Xu and P. Botchway and C. Brockett},
  title     = {Generating Role-Playing Game Quests with GPT Language Models},
  booktitle = {Proceedings of the IEEE Conference on Games},
  year      = {2023},
  doi       = {10.1109/CoG60054.2024.10645607}
}

@article{pangia,
author = {Buongiorno, Steph and Klinkert, Lawrence and Zhuang, Zixin and Chawla, Tanishq and Clark, Corey},
year = {2024},
month = {11},
pages = {156-166},
title = {PANGeA: Procedural Artificial Narrative Using Generative AI for Turn-Based, Role-Playing Video Games},
volume = {20},
journal = {Proceedings of the AAAI Conference on Artificial Intelligence and Interactive Digital Entertainment},
doi = {10.1609/aiide.v20i1.31876}
}

@misc{practical_pcg_llm,
      title={Practical PCG Through Large Language Models}, 
      author={Muhammad U Nasir and Julian Togelius},
      year={2023},
      eprint={2305.18243},
      archivePrefix={arXiv},
      primaryClass={cs.CL},
}

@article{pcg_llm_survey_2024,
  author  = {Mahdi Farrokhi Maleki and Richard Zhao},
  title   = {Procedural Content Generation in Games: A Survey with Insights on Emerging LLM Integration},
  journal = {Proceedings of the AAAI Conference on Artificial Intelligence},
  year    = {2024},
  note    = {arXiv:2410.15644}
}

@INPROCEEDINGS{arlpcg,
  author={Gisslén, Linus and Eakins, Andy and Gordillo, Camilo and Bergdahl, Joakim and Tollmar, Konrad},
  booktitle={2021 IEEE Conference on Games (CoG)}, 
  title={Adversarial Reinforcement Learning for Procedural Content Generation}, 
  year={2021},
  volume={},
  number={},
  pages={1-8},
  doi={10.1109/CoG52621.2021.9619053}}

@ARTICLE{pcgml_survey,
  author={Summerville, Adam and Snodgrass, Sam and Guzdial, Matthew and Holmgård, Christoffer and Hoover, Amy K. and Isaksen, Aaron and Nealen, Andy and Togelius, Julian},
  journal={IEEE Transactions on Games}, 
  title={Procedural Content Generation via Machine Learning (PCGML)}, 
  year={2018},
  volume={10},
  number={3},
  pages={257-270},
  doi={10.1109/TG.2018.2846639}}

@article{dungeon_pcg_survey,
  author  = {Roland van der Linden and Ricardo Lopes and Rafael Bidarra},
  title   = {Procedural Generation of Dungeons},
  journal = {IEEE Transactions on Computational Intelligence and AI in Games},
  volume  = {6},
  number  = {1},
  pages   = {78--89},
  year    = {2014},
  doi     = {10.1109/TCIAIG.2013.2297570}
}

@article{hendrikx_pcgsurvey,
  author  = {Mark J. C. Hendrikx and Sebastiaan A. Meijer and Joeri van der Velden and Alexandru Iosup},
  title   = {Procedural Content Generation for Games: A Survey},
  journal = {ACM Transactions on Multimedia Computing, Communications, and Applications},
  volume  = {9},
  number  = {1},
  year    = {2013},
  doi     = {10.1145/2422956.2422957}
}

@article{edpcg,
author = {Yannakakis, Georgios and Togelius, Julian},
year = {2011},
month = {07},
pages = {147-161},
title = {Experience-Driven Procedural Content Generation},
volume = {2},
journal = {Affective Computing, IEEE Transactions on},
doi = {10.1109/T-AFFC.2011.6}
}

@INPROCEEDINGS{lima_branching_quests,
  author={Soares de Lima, Edirlei and Feijó, Bruno and Furtado, Antonio L.},
  booktitle={2019 18th Brazilian Symposium on Computer Games and Digital Entertainment (SBGames)}, 
  title={Procedural Generation of Quests for Games Using Genetic Algorithms and Automated Planning}, 
  year={2019},
  volume={},
  number={},
  pages={144-153},
  doi={10.1109/SBGames.2019.00028}}

@misc{conan_quest_gen,
      title={Let CONAN tell you a story: Procedural quest generation}, 
      author={Vincent Breault and Sebastien Ouellet and Jim Davies},
      year={2018},
      eprint={1808.06217},
      archivePrefix={arXiv},
      primaryClass={cs.AI},
}

@INPROCEEDINGS{hu_llm_game_generation,
  author={Hu, Chengpeng and Zhao, Yunlong and Liu, Jialin},
  booktitle={2024 IEEE Conference on Games (CoG)}, 
  title={Game Generation via Large Language Models}, 
  year={2024},
  volume={},
  number={},
  pages={1-4},
  doi={10.1109/CoG60054.2024.10645597}}

@article{yuan_science_evaluation,
  author  = {Yuan, Jiayi and Zhang, Jiamu and Wen, Andrew and Hu, Xia},
  title   = {The Science of Evaluating Foundation Models},
  journal = {arXiv preprint},
  year    = {2025},
  eprint  = {2502.09670},
  archivePrefix = {arXiv},
  primaryClass  = {cs.CL}
}

@article{nasir2024word2world,
  title={Word2world: Generating stories and worlds through large language models},
  author={Nasir, Muhammad U and James, Steven and Togelius, Julian},
  journal={arXiv preprint arXiv:2405.06686},
  year={2024}
}

@article{alharthi2025generative,
  title={Generative AI in game design: Enhancing creativity or constraining innovation?},
  author={Alharthi, Sultan A},
  journal={Journal of Intelligence},
  volume={13},
  number={6},
  pages={60},
  year={2025},
  publisher={MDPI}
}

@article{de2022procedural,
  title={Procedural generation of branching quests for games},
  author={de Lima, Edirlei Soares and Feij{\'o}, Bruno and Furtado, Antonio L},
  journal={Entertainment Computing},
  volume={43},
  pages={100491},
  year={2022},
  publisher={Elsevier}
}

@inproceedings{papineni2002bleu,
  title={Bleu: a method for automatic evaluation of machine translation},
  author={Papineni, Kishore and Roukos, Salim and Ward, Todd and Zhu, Wei-Jing},
  booktitle={Proceedings of the 40th annual meeting of the Association for Computational Linguistics},
  pages={311--318},
  year={2002}
}

@inproceedings{lin2004rouge,
  title={Rouge: A package for automatic evaluation of summaries},
  author={Lin, Chin-Yew},
  booktitle={Text summarization branches out},
  pages={74--81},
  year={2004}
}

@inproceedings{novikova2017we,
  title={Why we need new evaluation metrics for NLG},
  author={Novikova, Jekaterina and Du{\v{s}}ek, Ond{\v{r}}ej and Curry, Amanda Cercas and Rieser, Verena},
  booktitle={Proceedings of the 2017 conference on empirical methods in natural language processing},
  pages={2241--2252},
  year={2017}
}

@article{celikyilmaz2020evaluation,
  title={Evaluation of text generation: A survey},
  author={Celikyilmaz, Asli and Clark, Elizabeth and Gao, Jianfeng},
  journal={arXiv preprint arXiv:2006.14799},
  year={2020}
}

@inproceedings{van2019best,
  title={Best practices for the human evaluation of automatically generated text},
  author={Van Der Lee, Chris and Gatt, Albert and Van Miltenburg, Emiel and Wubben, Sander and Krahmer, Emiel},
  booktitle={Proceedings of the 12th international conference on natural language generation},
  pages={355--368},
  year={2019}
}

@article{bommasani2021opportunities,
  title={On the opportunities and risks of foundation models},
  author={Bommasani, Rishi and Hudson, Drew A and Adeli, Ehsan and Altman, Russ and Arora, Simran and von Arx, Sydney and Bernstein, Michael S and Bohg, Jeannette and Bosselut, Antoine and Brunskill, Emma and others},
  journal={arXiv preprint arXiv:2108.07258},
  year={2021}
}

@article{liang2022holistic,
  title={Holistic evaluation of language models},
  author={Liang, Percy and Bommasani, Rishi and Lee, Tony and Tsipras, Dimitris and Soylu, Dilara and Yasunaga, Michihiro and Zhang, Yian and Narayanan, Deepak and Wu, Yuhuai and Kumar, Ananya and others},
  journal={arXiv preprint arXiv:2211.09110},
  year={2022}
}

\end{document}